\documentclass[11pt]{article}
\usepackage[utf8]{inputenc}
\usepackage[T1]{fontenc}
\usepackage[margin=1in]{geometry}
\usepackage{amsmath,amssymb}
\usepackage{booktabs}
\usepackage{graphicx}
\usepackage{xcolor}
\usepackage[hidelinks]{hyperref}
\usepackage{natbib}

\hypersetup{
  pdftitle={Self-Preference Is Weak or Absent in Verifiable Instruction-Following Revision},
  pdfauthor={William Guey, Pierrick Bougault}
}

\newcommand{\pp}{\,pp}
\newcommand{\CI}[2]{$[#1,\,#2]$}

\title{\textbf{Self-Preference Is Weak or Absent in Verifiable\\
Instruction-Following Revision: A Four-Model Test\\
Under Genuine Authorship}}

\author{
William Guey \quad Pierrick Bougault \\[3pt]
\normalsize Department of Industrial Engineering, Tsinghua University, Beijing, China \\[1pt]
\normalsize Correspondence: \texttt{guijt24@mails.tsinghua.edu.cn}
}
\date{June 2026}

\begin{document}
\sloppy
\maketitle

\begin{abstract}
\noindent
Large language models (LLMs) increasingly review and revise text, including their own.
A documented \emph{self-preference} bias (models favoring their own generations when
acting as judges) raises the question of whether models also resist valid corrections
to their own writing. We test this in a setting where ``valid'' is decided not by another
model but by a deterministic verifier: instruction-following revision on IFEval. A model
writes a draft; the official IFEval checker confirms the draft violates a constraint and
that a candidate edit fixes it; the model then accepts or rejects that edit either as the
genuine in-context \emph{author} or as a \emph{fresh} model that sees the draft neutrally.
Across four mid-tier model families and 85 author-versus-fresh comparisons, we find no
detectable self-preference: authors reject verified-good fixes to their own drafts at
essentially the same rate as fresh models judging the same drafts (gap $-5.1\pp$, 95\%
CI \CI{-12.9}{+2.7}). A self-skepticism hint from a smaller pilot did not replicate at
scale. The one robust observation is qualitative: when authors do reject a verified-good
fix, $97\%$ of their stated reasons are flaw-catching rather than preference, that is,
about the \emph{character} of rejections, not an elevated \emph{rate}. Effects smaller than
$\sim$13\pp{} cannot be excluded at this sample size.
\end{abstract}

\section{Introduction}

When a language model judges text, it tends to favor text it produced itself. This
\emph{self-preference} (or self-bias) is documented in the LLM-as-a-judge literature: models
both recognize and prefer their own generations \citep{panickssery2024llm}, and the bias
is amplified in self-refinement loops \citep{xu2024pride}. A natural and practically
important extension is \emph{revision}: when a model is shown a concrete, correct fix to a
draft it wrote, is it more reluctant to accept that fix than a model that did not write the
draft? If so, agentic ``self-review'' pipelines would be systematically handicapped on a
model's own output.

Answering this cleanly is hard for two reasons. First, ``is this fix an improvement?'' is
usually itself an LLM judgment, which reintroduces the very bias under study (circularity).
Second, the comparison must isolate \emph{authorship}: most prior probes use a told label
(``you wrote this'') rather than the model's genuine generation history.

We address both by restricting attention to \emph{verifiable} instruction following. Using
IFEval \citep{zhou2023instruction} and its official deterministic checker, ``the draft
violates constraint $c$'' and ``this edit makes the draft satisfy $c$ without breaking any
other constraint'' are ground-truth facts established by program, not by a model. We then
compare two deciders on each verified-good fix: the model that \emph{actually generated the
draft in-context} (the \emph{author}) versus a \emph{fresh} model that never saw the
generation history. The deterministic checker removes the circularity; the in-context
generation removes the told-label confound.

\paragraph{Contribution.} We test whether mid-tier LLMs treat verified-good fixes to their
\emph{own} instruction-following drafts differently than fixes to another model's draft,
under genuine in-context authorship, across four model families, and find \emph{no
detectable effect}. The deterministic-checker$+$author-vs-fresh design is the means that
makes this null credible; it is not claimed as a new method, and the result is an
\emph{absence in a previously unexamined setting}, not a confirmation of prior work. We
report three pilots honestly, including a self-skepticism hint that did \emph{not}
replicate, and one careful positive observation about the character (not rate) of author
self-rejections.

\section{Related Work}
\label{sec:related}

It is useful to read prior work as mapping different \emph{cells} of where self-related bias
does or does not appear, defined by the task the model performs on the text.

\textbf{Quality-judging.} Self-preference is clearest when a model scores or compares free-form
text quality. \citet{panickssery2024llm} show models recognize and favor their own generations,
with self-recognition correlated with the strength of the bias; \citet{xu2024pride} show
self-bias is amplified in self-refinement across translation, constrained generation, and
reasoning. These establish the phenomenon in the \emph{judging} cell, where the broader
LLM-as-judge paradigm \citep{zheng2023judging} is itself subject to a documented
\emph{self-enhancement} bias toward a model's own answers.

\textbf{Answer-selection.} When the task is selecting among candidate answers in multi-agent
settings, identity-driven effects appear weaker or are dominated by sycophancy, and can be
reduced by anonymization \citep{choi2026identity}. This suggests the bias is not uniform across
task framings.

\textbf{Feedback incorporation.} On verifiable tasks, models can resist even correct external
feedback \citep{jiang2025feedback}, and sycophancy under user rebuttal pulls in the opposite
direction \citep{kim2025challenging}. Crucially, this resistance is documented as a general
property, \emph{not} as something specific to a model's own authorship.

\textbf{Self-correction and session separation.} Self-refinement pipelines drive a model to
critique and revise its own output \citep{madaan2023selfrefine}, but surveys find no consensus
that LLMs reliably fix their own mistakes without an external signal \citep{kamoi2024when}, and
intrinsic self-correction without ground truth can even degrade reasoning \citep{huang2024large}.
Recent work studies separating production and review sessions to improve error-catching
\citep{song2026crosscontext}.

The cell we examine (\emph{verifiable revision under genuine in-context authorship}: does a
model accept a machine-verified-correct fix to its own draft less often than a fresh model?)
is, to our knowledge, unexamined. That is the gap this paper fills, with a null.

\section{Method}
\label{sec:method}

\paragraph{Substrate and verifier.} We use the real IFEval dataset \citep{zhou2023instruction}
(541 prompts, each with one or more verifiable instructions such as ``respond in all capital
letters'' or ``include the word X''). All pass/fail and fix labels are produced by IFEval's
\emph{official} checker (the \texttt{instruction\_following\_eval} instruction registry from
google-research), invoked through its strict-evaluation loop. No model is ever used to decide
whether a draft satisfies a constraint or whether a fix is correct.

\paragraph{Editable constraints.} Because we need fixes that a single minimal edit can make
pass, we restrict to a whitelist of \emph{atomically editable} instruction types: keyword
inclusion/exclusion and keyword frequency, all-caps or all-lowercase, a title in
double angle brackets, postscript, placeholder count, highlighted-section count, end-with-phrase,
quotation wrapping, bullet count ($\le 4$), and comma avoidance. Count-heavy constraints
(minimum sentence or word counts, large letter-frequency targets, paragraph counts) are
excluded because they have no clean atomic fix.

\paragraph{Drafts and one verified-good fix.} For each prompt, models generate drafts (normal
and a ``rushed'' system-prompt variant, to raise the natural failure rate). We keep drafts that
the checker says fail $\ge 1$ instruction. For each failing draft we generate targeted minimal
edits against the violated constraint's official natural-language description, score every
candidate with the checker, and keep \emph{exactly one} candidate that the checker labels
\textbf{GOOD}: it flips a failed constraint to pass and breaks no passing constraint. Using a
single fix per draft eliminates a redundancy artifact observed in an early pilot
(Section~\ref{sec:p1}). The fix labels GOOD, BAD, and NEUTRAL are defined purely by the checker
(BAD $=$ breaks a passing constraint; NEUTRAL $=$ no change in outcomes).

\paragraph{Author vs.\ fresh decision.} Each draft$+$fix is judged under two conditions.
\emph{Author (X):} the model that actually produced the draft decides ACCEPT/REJECT
\emph{inside its own conversation}, so the system prompt, the user task, its own draft as the
assistant turn, and then the edit question are all in context. This is genuine authorship, not
a told label. \emph{Fresh (Z):} a \emph{different} model, in a new conversation, sees the task,
the draft presented neutrally as ``a draft response,'' and the same edit, and answers the same
question. The decider is never told the IFEval constraints, that fixes are machine-verified, or
which fixes are good; it is asked to judge as an editor and to give a one-line reason. Each
condition is run three times at temperature $0.7$, with fix order irrelevant (one fix per
draft).

\paragraph{Measure.} For verified-good fixes define the \textbf{self-preference gap}
\[
G \;=\; r_{\text{author}} - r_{\text{fresh}},
\]
where $r$ is the rejection rate of the fix. $G>0$ means the author rejects valid fixes to its
own draft \emph{more} than a fresh model does (treating its own work more harshly, the
direction predicted both by self-preference or defense and by self-skepticism); $G<0$ means the
author is \emph{more} willing to accept corrections to its own work; $G\approx 0$ means no
authorship effect. The draft is the unit of analysis (rates are per-draft means over the three
reps); we report a 95\% CI from a paired bootstrap over drafts ($10{,}000$ resamples). We also
classify each author rejection's stated reason as \emph{stricter-than-checker} (catching a
flaw, e.g.\ ``this still is not fully uppercase'') versus \emph{preference} (``I prefer the
original''), to characterize \emph{why} authors reject when they do.

\section{Results}
\label{sec:results}

We ran three pilots of increasing rigor (Table~\ref{tab:summary}). All experiments use
OpenRouter-hosted mid-tier models, chosen deliberately because frontier models nearly ceiling
out IFEval and leave few fixable errors.

\begin{table}[t]
\centering
\small
\begin{tabular}{llrrr}
\toprule
Pilot & Authorship probe & $G$ (\pp) & 95\% CI (\pp) & $N$ \\
\midrule
P1: told label, 2 models     & ``a draft \emph{you} wrote''           & $-0.7$ & $\approx\pm9.5$         & 35 fixes \\
P2: genuine, 2 models        & in-context author vs.\ fresh           & $0.0$  & \CI{-13.6}{+13.6}      & 27 drafts \\
P3: genuine, 4 models        & in-context author vs.\ fresh           & $-5.1$ & \CI{-12.9}{+2.7}       & 85 drafts \\
\bottomrule
\end{tabular}
\caption{Self-preference gap $G=r_{\text{author/own}}-r_{\text{fresh/other}}$ (rejection-rate
difference on verified-good fixes). Positive $G$ would indicate self-preference. P1 is
expressed as a rejection gap for comparability (equivalently, acceptance of own-draft fixes was
$44.3\%$ vs.\ $43.5\%$). All three estimates are at or below zero with CIs spanning zero.}
\label{tab:summary}
\end{table}

\subsection{P1: told-label proxy (inconclusive, and why)}
\label{sec:p1}

A first pilot used a told label: a single decider judged the same draft$+$fixes framed once as
``a draft \emph{you} wrote'' and once as ``a draft written by \emph{another} AI model,''
byte-identical except one sentence. Pooled over two deciders (gpt-4o-mini,
claude-3.5-haiku), acceptance of verified-good fixes was $44.3\%$ (own) vs.\ $43.5\%$ (other),
i.e.\ $G=-0.7\pp$ ($\approx\pm9.5\pp$), with $35$ good fixes across $11$ drafts. Two problems
made this inconclusive. First, the manipulation is only a \emph{told} label on text the model
did not actually author in context. Second, we showed several near-duplicate good fixes per
draft, and roughly half of all rejections cited \emph{redundancy} (``already addressed by
another edit''), deflating and adding noise to acceptance. We also could not test specificity:
zero BAD fixes arose naturally, because atomic edits on independent constraints almost never
break a sibling constraint. P1 motivated the design changes in P2 and P3 (genuine authorship;
one fix per draft) and is reported only as the development path.

\subsection{P2: genuine authorship, two models (clean null)}

P2 fixes both problems: the author decides inside its own generation conversation, and exactly
one verified-good fix is shown per draft. Across $27$ drafts ($15$ gpt-authored, $12$
haiku-authored), authors and fresh models accepted good fixes at an \emph{identical} $84.0\%$,
so $G=0.0\pp$, 95\% CI \CI{-13.6}{+13.6}. The per-draft signs are symmetric ($4$ in the
self-preference direction, $4$ opposite, $19$ tied), with no directional lean under the mean.
The jump from P1's $\sim$44\% to $84\%$ acceptance confirms that the low P1 baseline was the
redundancy artifact, not authorship. The only residue was a weak asymmetry in the paired
fingerprint: $6$ drafts where the author rejected its own good fix while a fresh model accepted
it, versus $4$ in the mirror direction, a hint of mild \emph{self-skepticism} (authors a touch
harsher on their own work), which P3 was designed to power up and test for generality.

\subsection{P3: genuine authorship, four models (the powered test)}

P3 scales to $85$ usable drafts across four mid-tier families: \texttt{gpt-4o-mini},
\texttt{claude-3.5-haiku}, \texttt{gemini-2.5-flash-lite} (substituted for an unavailable
\texttt{gemini-flash-1.5}), and \texttt{llama-3.3-70b-instruct}. Each author is a draft's true
generator; the fresh decider is a different family, rotated so each family serves as both author
and comparator.

Pooled, authors rejected verified-good fixes to their own drafts $15.3\%$ of the time versus
$20.4\%$ for fresh models on the same drafts: $G=-5.1\pp$, 95\% CI \CI{-12.9}{+2.7}, which
includes zero. The self-skepticism hint from P2 did \emph{not} replicate; if anything the point
estimate is slightly negative (authors marginally \emph{more} willing to accept fixes to their
own work). The paired counts are balanced to mirror-leaning: $9$ author-rejects-fresh-accepts
versus $10$ in the mirror direction (any-rep).

\begin{table}[t]
\centering
\small
\begin{tabular}{lrrrrr}
\toprule
Author family & $N$ & author rej.\ & fresh rej.\ & $G$ (\pp) & 95\% CI (\pp) \\
\midrule
gpt-4o-mini            & 34 & $13.7\%$ & $8.8\%$  & $+4.9$  & \CI{-4.9}{+14.7} \\
gemini-2.5-flash-lite  & 19 & $17.5\%$ & $15.8\%$ & $+1.8$  & \CI{-17.5}{+19.3} \\
claude-3.5-haiku       & 19 & $15.8\%$ & $29.8\%$ & $-14.0$ & \CI{-29.8}{0.0} \\
llama-3.3-70b-instruct & 13 & $15.4\%$ & $43.6\%$ & $-28.2$ & \CI{-53.8}{-7.7} \\
\midrule
Pooled                 & 85 & $15.3\%$ & $20.4\%$ & $-5.1$  & \CI{-12.9}{+2.7} \\
\bottomrule
\end{tabular}
\caption{P3 per-author-family self-preference gap. Two families are positive, two negative;
there is no cross-family majority for either sign. Note that \emph{author} rejection rates are
tightly clustered ($13.7$ to $17.5\%$) while \emph{fresh} rejection rates vary widely
($8.8$ to $43.6\%$): the per-family gaps are driven mainly by how strict each author's
\emph{comparator} families happened to be, not by the authors themselves (Section~\ref{sec:disc},
limitation (c)).}
\label{tab:perfamily}
\end{table}

\paragraph{Cross-model (lack of) generality.} Table~\ref{tab:perfamily} shows two families with
small positive gaps (gpt, gemini) and two with negative gaps (haiku, llama); no sign holds in a
majority of families. The dispersion is almost entirely on the \emph{fresh} side: author
self-rejection sits near $15\%$ for all four families, whereas the comparator rejection rate
ranges from $8.8\%$ to $43.6\%$. The apparent per-family ``effects'' are therefore largely
comparator-strictness artifacts of the design (Section~\ref{sec:disc}).

\paragraph{What survives: the character of self-rejections.} Of the $39$ author self-rejections
of verified-good fixes, $38$ ($97.4\%$) were classified as \emph{stricter-than-checker}: the
author objects that the fix is inadequate or breaks something the verifier and fresh model let
through (e.g.\ ``the edit includes the name but breaks the limerick's AABBA scheme''). Only
$1$ was mere preference (Table~\ref{tab:reasons}). So \emph{when} an author rejects a fix to its
own work, it is almost never ``I prefer mine''; it is flaw-catching. This is a statement about
the \emph{character} of rejections, not about their \emph{rate}, which (Table~\ref{tab:perfamily})
is not elevated relative to fresh models.

\begin{table}[t]
\centering
\small
\begin{tabular}{lrr}
\toprule
Author-rejection reason class & Count & Share \\
\midrule
Stricter-than-checker (flaw-catching) & 38 & $97.4\%$ \\
Preference (``prefer the original'')  & 1  & $2.6\%$ \\
Other                                  & 0  & $0.0\%$ \\
\midrule
Total author self-rejections           & 39 & $100\%$ \\
\bottomrule
\end{tabular}
\caption{Reason classification for author self-rejections in P3 (classifier: gpt-4o-mini, fixed
rubric; all raw reasons released).}
\label{tab:reasons}
\end{table}

\section{Discussion and Limitations}
\label{sec:disc}

\paragraph{Reading across the three pilots.} A told-label proxy (P1), a clean two-model
authorship test (P2), and a powered four-model test (P3) all return self-preference gaps at or
below zero with CIs spanning zero. In verifiable instruction-following revision, mid-tier models
neither defend nor disproportionately distrust their own writing: they accept machine-verified
corrections to their own drafts at roughly the same high rate (about $80$ to $85\%$) as a fresh
model. Relating this to Section~\ref{sec:related}: self-preference is documented in
quality-\emph{judging} \citep{panickssery2024llm,xu2024pride}, appears weak or rare in
answer-\emph{selection} \citep{choi2026identity}, and resistance to correct feedback exists on
verifiable tasks but is not authorship-specific \citep{jiang2025feedback}. Our cell, verifiable
\emph{revision} under genuine authorship, shows no detectable authorship effect. The bias is
cell-dependent, and this cell appears clean.

\paragraph{Limitations.} \emph{(a) Mid-tier models only.} We deliberately used mid-tier models to
avoid IFEval's ceiling effect; frontier models are untested and could differ. \emph{(b) Low-latitude
constraints only.} IFEval constraints are atomic and machine-verifiable; holistic or high-latitude
revision (style, argumentation, factuality) is exactly where judging-cell self-preference is
strongest and is untested here. \emph{(c) Comparator confound.} In P2 and P3 the fresh decider is a
\emph{different} family with its own baseline strictness, so per-family gaps conflate authorship
with comparator strictness. This is visible in Table~\ref{tab:perfamily}: author rejection is flat
near $15\%$ across families while fresh rejection ranges $8.8\%$ to $43.6\%$, so the large negative
gaps for claude and llama reflect strict comparators, not lenient authors. A clean deconfounded
design would have \emph{all} non-author families judge each draft as fresh. \emph{(d) Power floor.}
With $N=85$ the pooled CI half-width is roughly $\pm 8$ to $\pm 13\pp$; effects smaller than
$\sim$13\pp{} cannot be excluded. We therefore claim ``weak or absent,'' never ``proven absent.''

\paragraph{The flaw-catching observation, stated carefully.} The one robust positive finding is
that author self-rejections are overwhelmingly flaw-catching ($97.4\%$ stricter-than-checker), not
preference-driven. This does \emph{not} say authors are net more self-critical; their rejection
\emph{rate} is not higher than fresh models'. It says that the rare self-rejections that do occur
are reasoned objections to the fix, not ego. We flag this as a hypothesis for future, properly
powered and deconfounded work, not as a demonstrated reversal; the self-skepticism rate signal from
P2 did not replicate in P3.

\section{Conclusion}

In verifiable, mid-tier instruction-following revision under genuine in-context authorship, we find
\emph{no detectable self-preference}: models accept machine-verified corrections to their own drafts
about as readily as a fresh model does (P3 gap $-5.1\pp$, 95\% CI \CI{-12.9}{+2.7}), with no
cross-family majority for either direction and effects below $\sim$13\pp{} unexcludable. A
self-skepticism hint from a smaller pilot did not replicate at scale. Placed against prior work, this
maps a clean cell: self-preference documented in quality-judging and weak in answer-selection is, in
this verifiable-revision setting, not detected. The lone durable observation is qualitative: when
authors do reject a verified-good fix, $97\%$ of reasons are flaw-catching rather than preference.
Future work should test frontier models, holistic (high-latitude) revision, and a fully deconfounded
all-families-as-fresh comparator before drawing rate-level conclusions.

\section*{Reproducibility}
All code, the official IFEval checker wrapper, fixed random seeds, and every raw decision (author and
fresh responses, one-line reasons, and checker labels) for all three pilots are released. The
deterministic checker means the verified-good and verified-bad labels are reproducible exactly;
model decisions are stochastic (temperature $0.7$, $3$ reps). See the repository \texttt{README}
for run order and the environment note (the dataset is loaded directly from the HuggingFace JSONL
to avoid a \texttt{datasets}/\texttt{fsspec} incompatibility). All code, data, and the full set of
raw decisions are available at \url{https://github.com/williamguey/self-preference-revision}.

\bibliographystyle{plainnat}
\bibliography{refs}

\end{document}